\documentclass[sigconf,screen]{acmart}

\usepackage{booktabs,tabularx,makecell,array}
\usepackage{wrapfig}
\usepackage[caption=false]{subfig}
\usepackage{siunitx}
\usepackage{microtype}

\setcopyright{cc}
\setcctype{by-nc-nd}
\acmDOI{10.1145/3776734.3794510}
\acmYear{2026}
\copyrightyear{2026}
\acmISBN{979-8-4007-2321-6/2026/03}
\acmConference[HRI Companion '26]{Companion Proceedings of the 21st ACM/IEEE International Conference on Human-Robot Interaction}{March 16--19, 2026}{Edinburgh, Scotland, UK}
\acmBooktitle{Companion Proceedings of the 21st ACM/IEEE International Conference on Human-Robot Interaction (HRI Companion '26), March 16--19, 2026, Edinburgh, Scotland, UK}
\received{2025-12-08}
\received[accepted]{2026-01-12}

\begin{document}

\title{Evolution 6.0: Robot Evolution through Generative Design}


\author{Muhammad Haris Khan}
\orcid{0009-0005-1112-0280}
\affiliation{%
  \institution{Skolkovo Institute of Science and Technology}
  \city{Moscow}
  \country{Russian Federation}}
\email{Haris.Khan@skoltech.ru}

\author{Artyom Myshlyaev}
\orcid{0009-0005-7449-7114}
\affiliation{%
  \institution{Skolkovo Institute of Science and Technology}
  \city{Moscow}
  \country{Russian Federation}}
\email{ Artyom.Myshlyaev@skoltech.ru}

\author{Artem Lykov}
\orcid{0000-0001-6119-2366}
\affiliation{%
  \institution{Skolkovo Institute of Science and Technology}
  \city{Moscow}
  \country{Russian Federation}}
\email{Artem.Lykov@skoltech.ru}

\author{Miguel Altamirano Cabrera}
\orcid{0000-0002-5974-9257}
\affiliation{%
  \institution{Skolkovo Institute of Science and Technology}
  \city{Moscow}
  \country{Russian Federation}}
\email{m.altamirano@skoltech.ru}

\author{Dzmitry Tsetserukou}
\orcid{0000-0001-8055-5345}
\affiliation{%
  \institution{Skolkovo Institute of Science and Technology}
  \city{Moscow}
  \country{Russian Federation}}
\email{d.tsetserukou@skoltech.ru}


\renewcommand{\shortauthors}{M. H. Khan, A. Myshlyaev, A. Lykov, M. Altamirano Cabrera, D. Tsetserukou}

\begin{abstract}

We propose a new concept, Evolution 6.0, which represents the evolution of robotics driven by Generative AI. When a robot lacks the necessary tools to accomplish a task requested by a human, it autonomously designs the required instruments and learns how to use them to achieve the goal. Evolution 6.0 is an autonomous robotic system powered by Vision-Language Models (VLMs), Vision-Language Action (VLA) models, and Text-to-3D generative models for tool design and task execution. The system comprises two key modules: the Tool Generation Module, which fabricates task-specific tools from visual and textual data, and the Action Generation Module, which converts natural language instructions into robotic actions. It integrates QwenVLM for environmental understanding, OpenVLA for task execution, and Llama-Mesh for 3D tool generation. Evaluation results demonstrate a 90\% success rate for tool generation with a 10-second inference time and action generation achieving 83.5\% in physical and visual generalization, 70\% in motion generalization, and 37\% in semantic generalization. Future improvements will focus on bimanual manipulation, expanded task capabilities, and enhanced environmental interpretation to improve real-world adaptability.
\end{abstract}


\maketitle

\section{Introduction}
The rapid development of generative AI has unlocked extraordinary opportunities for robotics automation~\cite{xu2024surveyroboticsfoundationmodels}, \cite{11128224}. Visual Language Action (VLA) models enable cognitive robots to perform complex manipulation tasks by leveraging instructions and visual information to generate control signals, bridging perception and action. Industry 6.0~\cite{lykov2024industry} envisions heterogeneous robot swarms autonomously designing, manufacturing, and assembling products based on user specifications, eliminating human involvement and enabling autonomous tool fabrication. However, Industry 6.0 cannot assess its environment or autonomously determine required tools, relying on direct user instructions that limit adaptability in unforeseen challenges.
Unlike Industry 6.0's predefined industrial settings, we propose Evolution 6.0, extending these capabilities to unpredictable environments like Mars, where tool scarcity and environmental variability pose major challenges. Evolution 6.0 empowers robots to analyze unfamiliar surroundings, interpret contextual cues via VLM, and generate solutions with VLA models. This enables autonomous tool determination, design, and fabrication, eliminating dependency on predefined instructions and enhancing operational flexibility.

\begin{figure*}[t]
    \centering

    \begin{minipage}{0.75\textwidth}
        \centering
        \includegraphics[width=\textwidth, alt={Figure shows the system architecture of the tool generation, which has as output the 3D model of the generated object that is 3D printed in the top and the action generation for the movement of the robot.}]{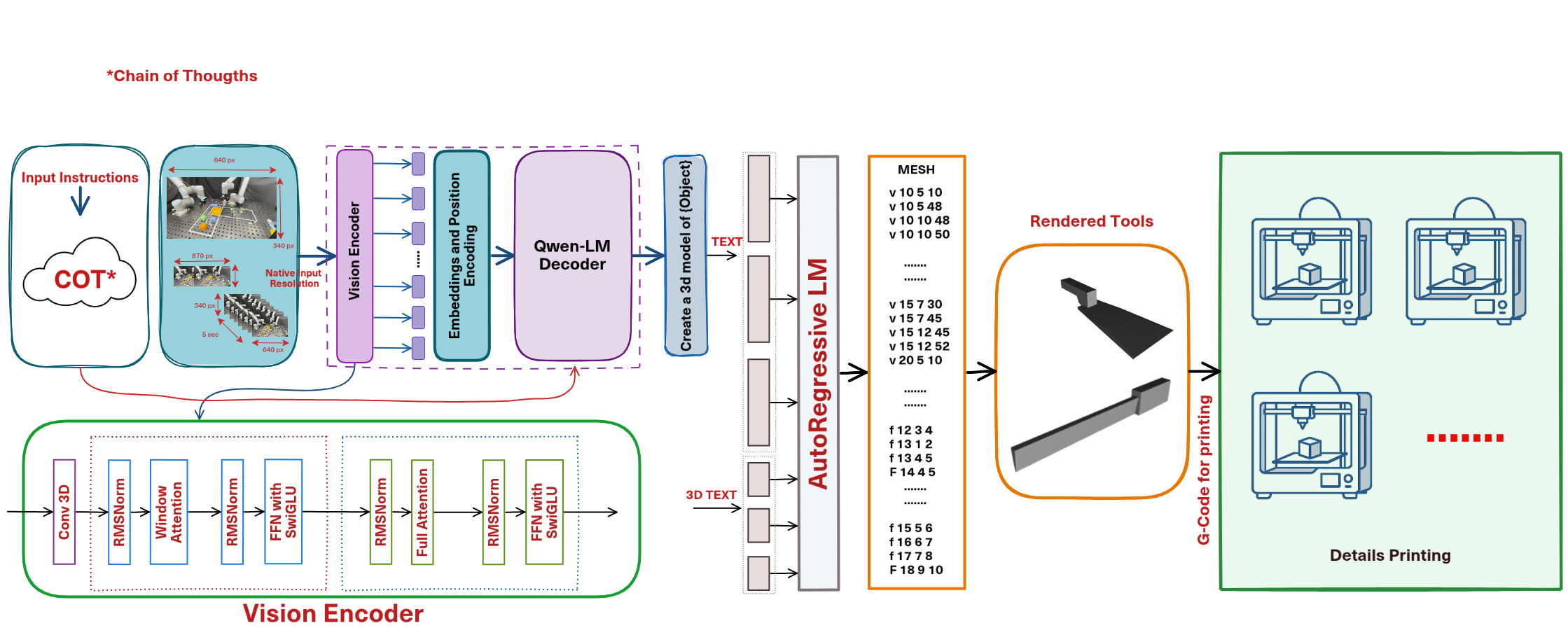}
        
    \end{minipage}

    \vspace{1.2em} 

    \begin{minipage}{0.75\textwidth}
        \centering
        \includegraphics[width=\textwidth, alt={Figure shows the system architecture of the tool generation, which has as output the 3D model of the generated object that is 3D printed in the top and the action generation for the movement of the robot.}]{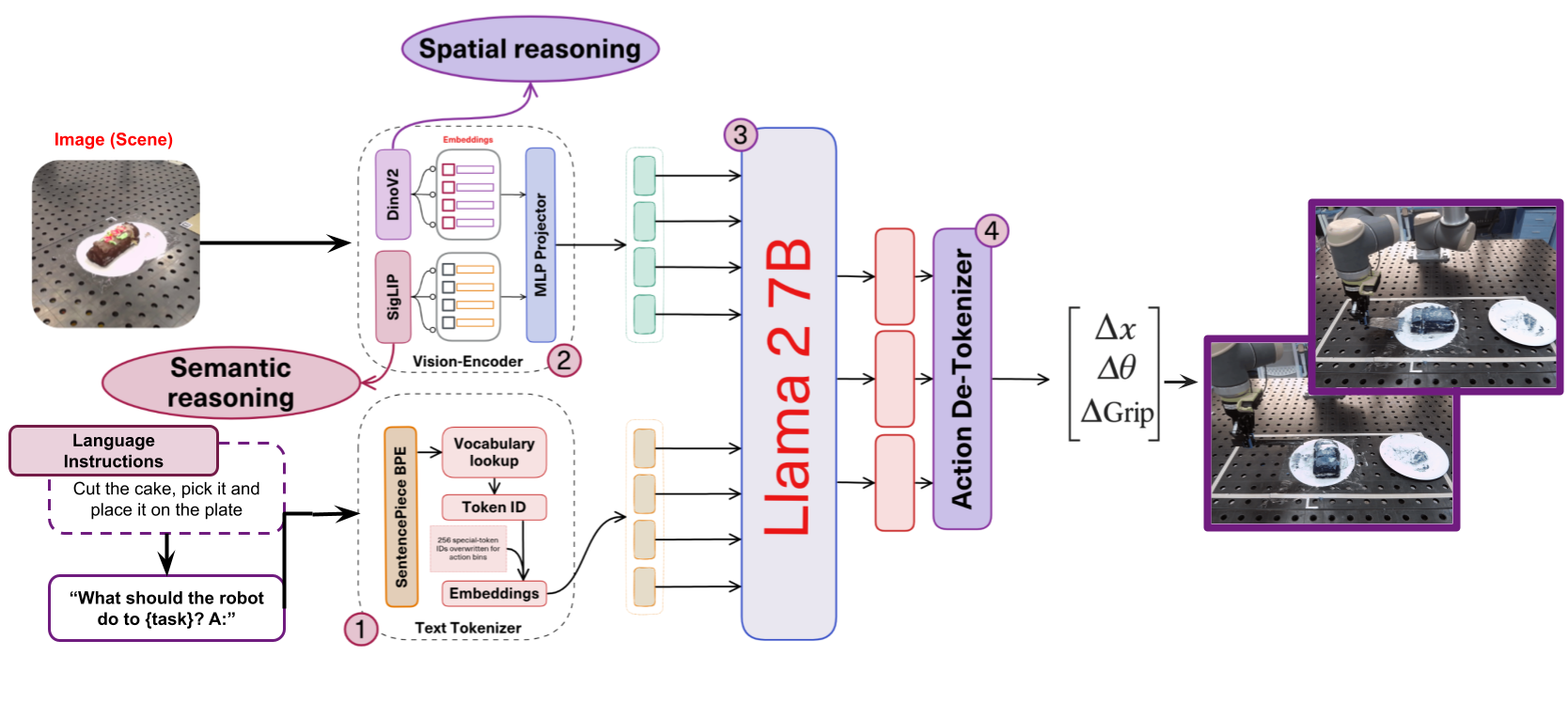}
        
    \end{minipage}

    \caption{Overview of the system architecture showing tool generation (top) followed by action generation (bottom).}
    \label{fig:arch}
\end{figure*}

Inspired by early humans crafting tools to overcome challenges, Evolution 6.0 enables robots to autonomously perceive surroundings, conceptualize solutions, and fabricate tools in real time. By integrating VLA models with Text-to-3D generative models, robots analyze novel challenges, dynamically design task-specific tools, and apply them seamlessly. In remote environments, where predefined toolsets are impractical, this approach offers transformative potential for responding to unforeseen challenges with high autonomy. Evolution 6.0 is inherently designed to generalize across open-ended tasks by dynamically interpreting new scenarios and autonomously generating required tools. The presented evaluation serves as a benchmark illustrating system effectiveness without implying capability limitations. Evolution 6.0 represents a paradigm shift, fostering adaptability where traditional manufacturing infrastructures are unavailable and paving the way for self-sufficient, adaptable robotic systems.
\begin{figure*}[!t]
  \centering
  \includegraphics[width=0.75\textwidth, alt={Figure describes the generation of the CAD models, it shows four examples of generated tool, having in the top of each one the prompt used. }]{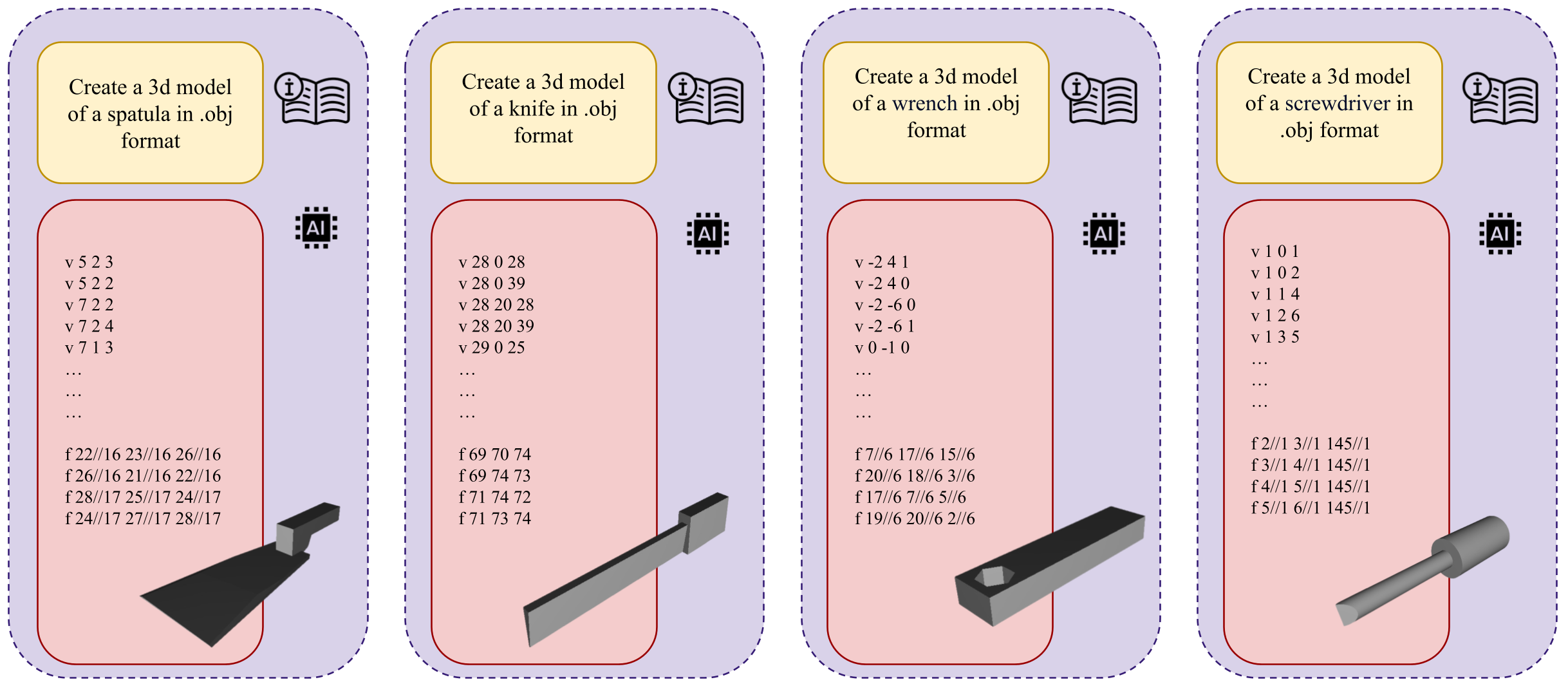}
  \caption{CAD model design by Generative AI.}
  \label{fig:cad}
\end{figure*}
\section{Related Work}

The development of autonomous systems for tool design, fabrication, and utilization hinges on advancements in Vision-Language Models (VLMs), Vision-Language Action (VLA) models, and Text-to-3D models. VLMs like BLIP-2~\cite{li2023blip}, Flamingo~\cite{alayrac2022flamingo}, Kosmos-2~\cite{peng2023kosmos}, Molmo and pixmo~\cite{deitke2024molmo}, and Qwen2-VL~\cite{wang2024qwen2} enable robots to interpret multimodal inputs for informed decision-making. VLA models generate precise 7D action outputs from natural language and visual data~\cite{gbagbe2024bi},\cite{khan2025shake}, with pioneering work including PaLM-E\cite{driess2023palm} and the RT-series (RT-1~\cite{brohan2022rt}, RT-2~\cite{brohan2023rt}, RT-X~\cite{o2023open}), alongside OpenVLA~\cite{kim24openvla} and MiniVLA~\cite{belkhale2024minivla}. Text-to-3D models~\cite{liu2022towards},\cite{jain2022zero},\cite{yu2023text},\cite{liu2024sherpa3d} convert language into 3D geometries, evolving from Text2Shape\cite{chen2019text2shape} and ShapeCrafter~\cite{fu2022shapecrafter} to GPT-4o-based SDF generation~\cite{openai2024gpt4technicalreport}\cite{osher2004level}, with efficient alternatives like LLaMa-Mesh\cite{wang2024llama} and MeshGPT~\cite{siddiqui2024meshgpt}. Our work integrates these technologies into Evolution 6.0, establishing a comprehensive framework for autonomous tool design and deployment.

\section{System Architecture of Evolution 6.0}

\subsection{Tool Generation Module}

The Tool Generation Module analyzes the robot's environment and autonomously generates necessary tools for task completion. It begins with Robot Scene Input, where sensors like cameras capture the environment to identify objects requiring interaction. Captured data is processed through Qwen2-VL-2B-Instruct, where a Vision Encoder extracts visual features interpreted by the QwenLM decoder to generate contextual textual descriptions. Based on scene analysis, the system formulates prompts such as ``Create a 3D model of a knife,'' passed to an Autoregressive Language Model (LlamMa-Mesh). This model synthesizes a 3D tool representation in mesh format containing vertex (v) and face (f) data defining the tool's geometry. The generated mesh undergoes Mesh Rendering for visualization and validation to ensure task requirements are met. Once validated, the finalized mesh converts to G-Code for 3D printer fabrication. The system scales tools by a factor depending on the manipulated item's size in the scene. Fig. \ref{fig:cad} shows tool examples produced by LlamMa-Mesh.

Evolution 6.0, illustrated in Fig. \ref{fig:arch}, integrates two connected modules enabling flexible robotic manipulation. The Tool Generation Module creates situation-specific tools, while the Action Generation Module determines task execution. This design adapts actions based on environmental observations and natural language instructions. The approach uses QwenVLM for interpreting surroundings, OpenVLA for determining task steps, and LlamMa-Mesh for shaping three-dimensional tools, enabling robots to operate effectively in environments requiring tailored solutions and adaptive behaviors. Overall, the Tool Generation Module provides a dynamic, adaptive mechanism for equipping robots with instruments needed to perform tasks in ever-changing environments.
\subsection{Action Generation Module}

The Action Generation Module converts natural language instructions into precise robotic actions using a multimodal approach. Built on OpenVLA \cite{kim24openvla} from Stanford AI Lab, it was fine-tuned on cake-cutting and cake-picking tasks, with each episode corresponding to a remotely controlled robotic manipulator task for effective adaptation to dynamic environments. The Vision-Language-Action (VLA) architecture takes third-person camera frames and natural language descriptions as input, outputting a 7D action vector controlling the manipulator through spatial displacements \( \Delta X \), angular movements \( \Delta \theta \), and gripping actions \( \Delta Grip \).

The system operates iteratively, continuously processing environmental input for real-time action adjustments. Upon receiving a task description, it analyzes visual data and textual instructions to compute and execute actions continuously, ensuring smooth, uninterrupted task execution by eliminating delays between movements. This enables real-time environmental response, improving task efficiency and adaptability.

For real-time processing, QwenVLM and LlamMa-Mesh models were optimized to int8 precision, while the fine-tuned VLA model retained original precision. This optimization, combined with an NVIDIA RTX 4090 24Gb GPU and streamlined communication protocols, enables \SI{5}{Hz} operation, balancing computational efficiency and responsiveness for Evolution 6.0's complex manipulation tasks.

\section{Dataset Collection and Training Pipeline}

The dataset was collected using a UR10 robotic arm with a Logitech C920e HD camera, comprising 20 episodes across cutting cake and pick-and-place tasks. Each episode recorded gripper position, orientation, state, visual feedback, and language instructions like ``Cut one piece of cake''. Data was formatted into RLDS structure for OpenVLA compatibility. The dataset fine-tuned an OpenVLA-7b model using parameter-efficient LoRA with rank-32 adapters. Training used a batch size of 16, a learning rate $5 \times 10^{-4}$, and 4000 gradient steps on a single A100 GPU. Training loss, action accuracy, and L1 loss were monitored for stable progress tracking.

\begin{table*}[t]
\centering
\vspace{3mm} 
\caption{Phase one results for tool generation, showing success rates and average inference times.}
\label{table:tool_generation_results}
\begin{tabular}{|c|c|c|}
\hline
\textbf{Task} & \textbf{Success Rate} & \textbf{Average Inference Time (seconds)} \\
\hline
Environmental interpretation via VLM & 80\% & 4 \\
3D tool generation & 90\% & 10 \\
\hline
\end{tabular}
\end{table*}

\begin{figure*}[!t]
  \centering
  \includegraphics[width=0.82\textwidth, alt={This plot describes the success rate across generalization categories: Semantic (41\% for Task 1, 33\% for Task 2 and 37\% in Average), Visual (83\% for Task 1, 84\% for Task 2 and 83.5\% in Average), Motion (75\% for Task 1, 65\% for Task 2 and 70\% in Average), and Physical (84\% for Task 1, 80\% for Task 2 and 83.5\% in Average). }]{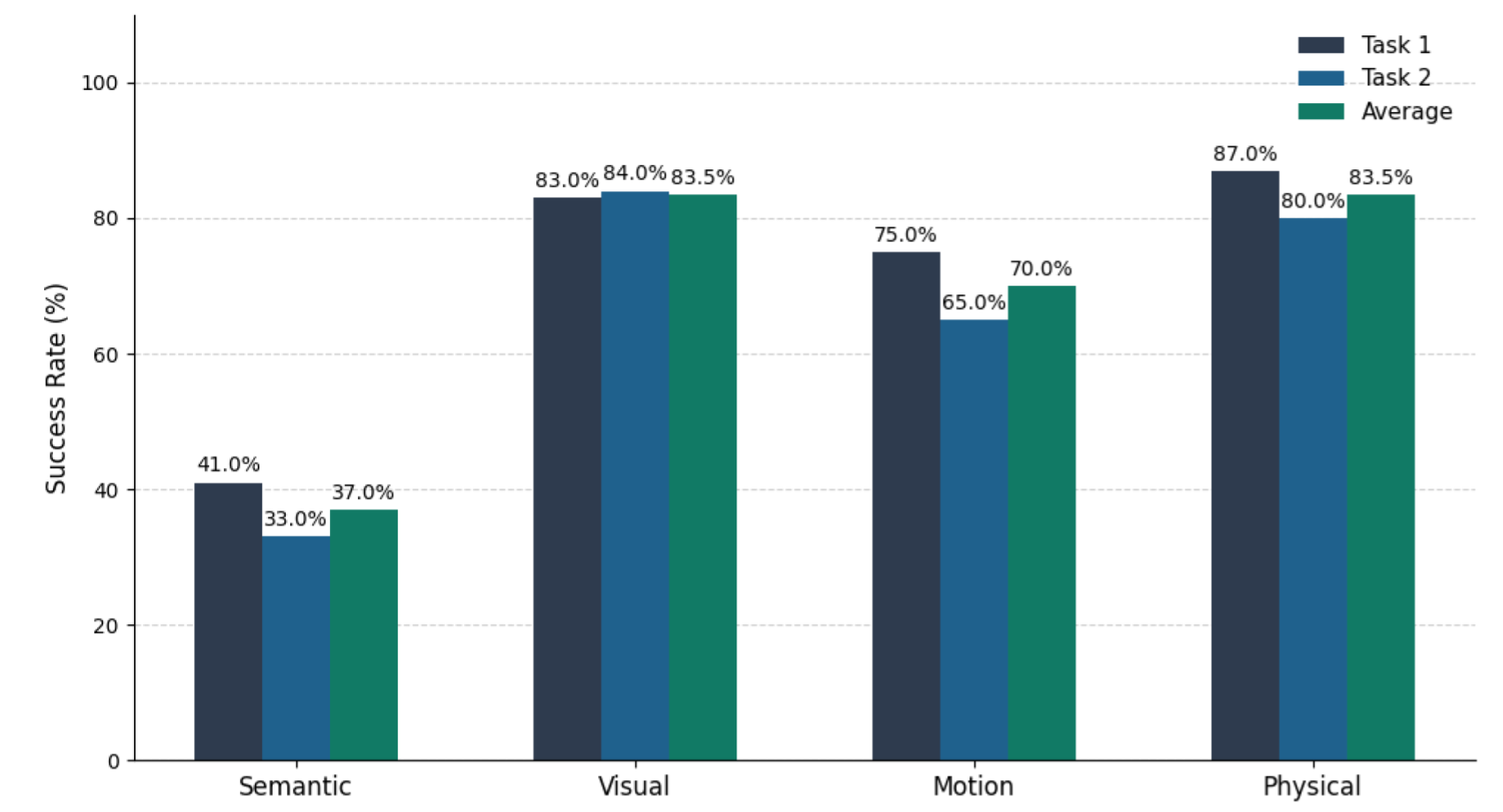}
  \caption{Phase two results: success rates across generalization categories.}
  \label{fig:res}
\end{figure*}

\section{Experiments}
\subsection*{Evaluation Methodology}
The evaluation consisted of two phases. First, the tool generation module was assessed using success rate and inference time. Second, the action generation module was evaluated on a UR-10 robot from a third-person view, with success rate as the primary metric across five scenarios: the first provided previously encountered training instructions and scenes, while four tested generalization abilities—physical (variations in object size/color), motion (changes in object position), semantic (new instructions like replacing ``cut the cake'' with ``cut the banana''), and visual (altered scenes or distractors). Task 1 and Task 2 validated system feasibility rather than limitations, with Evolution 6.0's architecture supporting seamless adaptation to new tasks without retraining through robust multimodal integration.

\subsubsection{Phase One: Tool Generation}

The first phase evaluated the tool generation module based on success rate and inference time. The Qwen2-VL-2B-Instruct model interprets environmental contexts and generates instructions for an autoregressive language model (LM) to create tools. Tested across 10 robotic scenarios, the VLM accurately described 8, producing appropriate prompts for the Text-to-3D model with an average inference time of \SI{4}{s}. Challenges arose with scenes containing multiple identical objects or same-class targets. For instance, when distinguishing screws from bolts, it misinterpreted the prompt as ``Create the 3D model of the screwdriver,'' incompatible with bolts. Despite this, the autoregressive LM achieved high success, generating tools in 9 of 10 instances with an average generation time of 10 seconds, varying by complexity. However, the model struggled with complex curves or fillets; as shown in Fig. \ref{fig:cad}, generated tools predominantly exhibited sharp edges or simple circular cross-sections, highlighting room for improvement in handling intricate geometrical features. Results are summarized in Table 1.

\subsubsection{Phase Two: Action Generation}

In the second phase, the action generation module was evaluated using a UR-10 robotic arm, with the success rate serving as the primary evaluation metric, following \cite{kim24openvla}. Testing spanned 10 distinct scenarios across physical, motion, visual, and semantic generalization to assess adaptability and robustness.
For Task 1, the module achieved success rates of 87\% in physical generalization, 83\% in visual generalization, 75\% in motion generalization, and 41\% in semantic generalization. 
In Task 2, performance slightly declined with success rates of 80\% in physical generalization, 84\% in visual generalization, 65\% in motion generalization, and 33\% in semantic generalization.
Results (summarized in Fig. \ref{fig:res}) highlight proficiency in physical and visual generalization and resilience under varying conditions. However, significant improvement is needed in motion and semantic generalization. For instance, instructions like ``Cut the banana'' or ``Place the tomato on the plate'' exhibited inconsistent behavior.
Overall, while the action generation module shows promise in familiar settings and certain generalizations, further refinement is needed to enhance reliability in complex or novel scenarios.
\section{Conclusion and Future Work}

Evolution 6.0 is a novel framework integrating Vision-Language Models (VLMs), Vision-Language Action (VLA) models, and Text-to-3D generative models to enable autonomous robotic tool design and utilization in dynamic environments. Experimental results demonstrate system effectiveness, with the tool generation module achieving a 90\% success rate and 10-second average inference time, while the action generation module reached average success rates of 83.5\% in physical and visual generalization, 70\% in motion generalization, and 37\% in semantic generalization. These outcomes highlight adaptability and real-world potential, while indicating improvement areas in semantic generalization and complex geometric tool generation. Future work will extend the framework to bimanual manipulation for coordinated dual-arm control in tasks like cooperative object handling and assembly, incorporate expanded task sets to enhance versatility in precision-demanding industrial environments, and optimize environmental interpretation by integrating advanced models and imitation learning techniques to improve accuracy in complex, cluttered, or ambiguous scenarios.

\section*{Acknowledgements} 
Research reported in this publication was financially supported by the RSF grant No. 24-41-02039.

\balance


\end{document}